\documentclass[runningheads]{llncs}
\usepackage[T1]{fontenc}
\usepackage{graphicx}
\usepackage{multirow}
\usepackage{hyperref}
\usepackage{todonotes}
\usepackage{comment}

\begin{document}
\title{An Investigation Into Race Bias in Random Forest Models Based on Breast DCE-MRI Derived Radiomics Features %\thanks{Supported by organization x.}
}

\titlerunning{Race Bias in Breast DCE-MRI Radiomics Features}

\author{Mohamed Huti\inst{1}%\orcidID{0000-1111-2222-3333}
\and
Tiarna Lee\inst{1}%\orcidID{1111-2222-3333-4444}
\and
Elinor Sawyer\inst{2}%\orcidID{2222--3333-4444-5555}
 \and
Andrew P. King\inst{1}%\orcidID{0000-0002-9965-7015}
}
\authorrunning{M. Huti et al.}

\institute{School of Biomedical Engineering and Imaging Sciences, King's College London, UK \and
School of Cancer and Pharmaceutical Sciences, King’s College London, UK}

%\author{**}
%\institute{**}
%\authorrunning{** et al.}


%\email{a@b.c}

%
\maketitle              % typeset the header of the contribution
\begin{abstract}
%The abstract should briefly summarize the contents of the paper in
Recent research has shown that artificial intelligence (AI) models can exhibit bias in performance when trained using data that are imbalanced by protected attribute(s). Most work to date has focused on deep learning models, but classical AI techniques that make use of hand-crafted features may also be susceptible to such bias. In this paper we investigate the potential for race bias in random forest (RF) models trained using radiomics features. Our application is prediction of tumour molecular subtype from dynamic contrast enhanced magnetic resonance imaging (DCE-MRI) of breast cancer patients. Our results show that radiomics features derived from DCE-MRI data do contain race-identifiable information, and that RF models can be trained to predict White and Black race from these data with 60-70\% accuracy, depending on the subset of features used. Furthermore, RF models trained to predict tumour molecular subtype using race-imbalanced data seem to produce biased behaviour, exhibiting better performance on test data from the race on which they were trained.

\end{abstract}
\section{Introduction}

The potential for artificial intelligence (AI) models to exhibit bias, or disparate performance for different protected groups, has been demonstrated in a range of computer vision and more recently medical imaging applications. For example, biased performance has been reported in AI models for diagnostic tasks from chest X-rays \cite{Larrazabal2020,Seyyed-Kalantari2021}, cardiac magnetic resonance (MR) image segmentation \cite{PuyolAnton2021,PuyolAnton2022,Lee2022}, brain MR image analysis \cite{Ioannou2022,Petersen2022,Wang2023,Stanley2022} and dermatology image analysis \cite{Abbasi-Sureshjani2020,Guo2021}. In response, the field of \emph{Fair AI} has emerged to address the challenge of making AI more trustworthy and equitable in its performance for protected groups \cite{Mehrabi2021}.

A common cause of bias in AI model performance is the combination of a distributional shift between the data of different protected groups and demographic imbalance in the training set. For example, in chest X-rays there is a distributional shift between sexes due to the presence of breast tissue lowering the signal-to-noise ratio of images acquired from female subjects \cite{Larrazabal2020}. However, more subtle distributional shifts can also exist which cannot be perceived by human experts, and recent work has shown that race-based distributional shifts are present in a range of medical imaging modalities, including breast mammography \cite{Gichoya2022}. This raises the possibility of race bias in AI models trained using imbalanced data from these modalities.

Most work on AI bias to date has focused on deep learning techniques, in which the features used for the target task are optimised as part of the training process. In the presence of distributional shift and training set imbalance this learning process can lead to bias in the features and potentially in model performance. Classical AI approaches are trained using fixed hand-crafted features such as radiomics, and so might be considered to be less susceptible to bias. However, despite these approaches still being widely applied, little experimental work has been performed to assess the potential for, and presence of, bias in these features and the resulting models.

In this paper, we investigate the potential for bias in a classical AI model (Random Forest) based on radiomics features. Our chosen application is potential race bias in Random Forest models trained using radiomics features derived from dynamic contrast enhanced magnetic resonance imaging (DCE-MRI) of breast cancer patients. This application is of interest because there have been reported differences in breast density and composition between races \cite{McCarthy2016,Moore2020}, as well as tumour biology \cite{Martini2022}, indicating a possible distributional shift in (imaging) data acquired from different races, and hence the possibility of bias in AI models trained using these data. Our target task is the prediction of tumour molecular subtype from the radiomics features. This is a clinically useful task because different types of tumour are commonly treated in different ways (e.g. surgery, chemotherapy), and tumour molecular subtype is normally determined through an invasive biopsy. Therefore, development and validation of an AI model to perform this task from imaging data would obviate the need for such biopsies.

This paper makes two key contributions to the field of Fair AI. First, we present the first thorough investigation into possible bias in AI models based on radiomics features. Second, we perform the first investigation of bias in AI models based on features derived from breast DCE-MRI imaging.

\section{Materials}
\label{sect:materials}

In our experiments we employ the dataset described in \cite{Saha2018}\footnote{The dataset is publicly available at: \href{https://doi.org/10.7937/TCIA.e3sv-re93}{https://doi.org/10.7937/TCIA.e3sv-re93}.}, which features pre-operative DCE-MRI images acquired from 922 female patients with invasive breast cancer at Duke Hospital, USA, together with demographic, clinical, pathology, genomic, treatment, outcome and other data. From the DCE-MRI images, 529 radiomics features have been derived which are split into three (partially overlapping) categories: whole breast, fibroglandular tissue (FGT) only and tumour only. The full dataset consists of approximately 70\% White subjects, 22\% Black subjects and 8\% other races. We refer the reader to \cite{Saha2018} for a full summary of patient characteristics and the data provided.

\section{Methods}

For all experiments we employed a Random Forest (RF) classifier as our AI model, similar to the work described in \cite{Saha2018}.
For each model, we performed a grid search hyperparameter optimisation using a 5-fold cross validation on the training set. Following this, the final model was trained with the selected hyperparameter values using all training data and applied to the test set. The hyperparameters optimised were the number of trees (50, 100, 200, 250), the maximum depth of the trees (10, 15, 30, 45) and the splitting criterion (entropy, Gini).
Our model training differed from that described in \cite{Saha2018} in three important ways:
\begin{enumerate}
    \item We used only Black and White subjects to enable us to analyse bias in a controlled environment. Data from all other races were excluded from both the training and test sets. This meant that our dataset comprised 854 subjects (651 White and 203 Black).
    \item To simplify our analysis, we focused on just one of the binary classification problems reported in \cite{Saha2018}: prediction of \emph{Luminal A} vs \emph{non-Luminal A} tumour molecular subtype. Based on these labels, the numbers of positive (\emph{Luminal A}) and negative (\emph{non-Luminal A}) subjects for each race are summarised in Table \ref{table:labels}. As can be seen, there is a higher proportion of \emph{non-Luminal A} tumours in the Black patients, which is consistent with prior studies on relative incidence of tumour subtypes by race \cite{AbdElRehim2004,Jones2022}.
    \item We did not perform feature selection prior to training and evaluating the RF classifiers. We chose to omit this step because one of our objectives was to analyse which specific radiomics features (if any) could lead to bias in the trained models, so we did not want to exclude any features prior to this analysis.
\end{enumerate}

\begin{table}[ht]
\begin{center}
\begin{tabular}{|l|p{1.5cm}|p{1.5cm}|p{1.5cm}|}
\hline
\textbf{Label} & \multicolumn{1}{|c|}{\textbf{White}} & \multicolumn{1}{|c|}{\textbf{Black}} & \multicolumn{1}{|c|}{\textbf{~~All~~~}} \\ \hline
Positive (\emph{Luminal A}) & \multicolumn{1}{|c|}{442} & \multicolumn{1}{|c|}{107} & \multicolumn{1}{|c|}{549} \\ \hline
Negative (\emph{non-Luminal A}) & \multicolumn{1}{|c|}{209} & \multicolumn{1}{|c|}{96} & \multicolumn{1}{|c|}{305} \\ \hline
\end{tabular}
\caption{Summary of positive (\emph{Luminal A}) and negative (\emph{non-Luminal A}) labels in the dataset overall and broken down by race.}
\label{table:labels}
\end{center}
\end{table}

\section{Experiments and Results}

\subsection{Race Classification}
\label{section:raceclassification}

In the first experiment, our aim was to determine if the radiomics features contain race-identifiable information. The presence of such information is a known potential cause of bias in trained models as it would be indicative of a distributional shift in the data between races, not just in the imaging data but in the derived (hand-crafted) radiomics features. To investigate this, we trained RF classifiers to predict race (White or Black) from the entire radiomics feature set, and also for the whole breast, FGT and tumour features individually. For these experiments, to eliminate the effect of class (i.e. race) imbalance, we randomly sampled from the dataset to create race-balanced training and test sets, each consisting of 100/100 White/Black subjects.

Results are reported as percentage classification accuracy in Table \ref{table:raceclass} for all subjects in the test set and also separately for each race. We can see that it is possible to predict race from radiomics features with around 60-70\% accuracy. The results are similar for both White and Black subjects and do not differ significantly for the category of radiomics features used. It should be noted that the whole breast, FGT and tumour categories are partially overlapping, hence the similar performance for the different radiomics categories. Specifically, a set of features related to breast and FGT volume is included in both the whole breast and FGT categories, and another set related to FGT and tumour enhancement is present in both the FGT and tumour categories \cite{Saha2018}.

\begin{table}[ht]
\begin{center}
\begin{tabular}{|l|c|c|c|}
\hline
\textbf{Radiomics features} & \textbf{Whole test set} & \textbf{White subjects only} &  \textbf{Black subjects only}\\ \hline
All & 63\% & 64\% & 66\% \\ \hline
Whole breast only & 62\% & 70\% & 57\% \\ \hline
FGT only & 61\% & 65\% & 60\% \\ \hline
Tumour only & 62\% & 62\% & 66\% \\ \hline
\end{tabular}
\caption{Race classification accuracy from radiomics features derived from breast DCE-MRI. Results are presented as percentage classification accuracy and reported for whole test set as well as broken down by race. Classification was performed from all radiomics features as well as just those derived from the whole breast, fibroglandular tissue (FGT) and tumour only.}
\label{table:raceclass}
\end{center}
\end{table}

\subsection{Bias Analysis}
\label{BiasExpt}

Having established one of the key conditions for the presence of bias in AI models, i.e. a distributional shift between the data of different protected groups, we next investigated whether training with highly imbalanced training sets can lead to bias in performance.

For these experiments we split the dataset into a training set of 426 subjects and a test set of 428 subjects. The split was random under the constraints that the White and Black subjects and the \emph{Luminal A} and \emph{non-Luminal A} subjects were evenly distributed between train and test sets.
The training set consisted of 325/101 White/Black subjects and 274/152 \emph{Luminal A}/\emph{non-Luminal A} subjects, and the test set consisted of 326/102 White/Black subjects  and 275/153 \emph{Luminal A}/\emph{non-Luminal A} subjects.

In addition, we curated two additional training sets consisting of only the White subjects and only the Black subjects from the combined training set described above. Due to the racial imbalance in the database, these training sets consisted of 325/101 subjects for White/Black subjects. Using all three training sets (i.e. all, White-only and Black-only), we trained RF models for the task of classifying \emph{Luminal A} vs \emph{non-Luminal A} tumour molecular subtype and evaluated their performance for the entire test set as well as for the White subjects and the Black subjects in the test set individually.
Class (i.e. molecular subtype) imbalance was addressed by applying a weighting to training samples that was inversely proportional to the class frequency.

Results are presented in Table \ref{table:luminalclass}, in which performance is quantified using the percentage classification accuracy. We performed this experiment using all radiomics features, just the whole breast features, just the FGT features and just the tumour features. We can see that in terms of overall performance, the models trained using all data and the White-only data had higher accuracy than the models trained using Black-only data, reflecting the impact of different training set sizes. Regarding race-specific performance, the models trained using all training data (i.e. 325/101 White/Black subjects) performed slightly better on White subjects, likely reflecting the effect of training set imbalance. The difference in performance in favour of White subjects varied from 3-11\% (mean 6.25\%), depending on the subset of features used. The models trained using White-only data had a larger performance disparity in favour of White subjects, varying between 6-11\% (mean 9\%). The models trained using Black-only data showed generally better performance on Black subjects (mean 3.5\% difference), although the model trained using all radiomics features was 1\% better for White subjects. In contrast, the model trained using whole breast radiomics features performed 10\% better for Black subjects. With the exception of this last result, in general there was not a noticeable difference in bias between the models trained using all radiomics features, just whole breast features, just FGT features and just tumour features, which is consistent with the similar race classification results reported in Section \ref{section:raceclassification}.

\begin{table}[]
\begin{center}
\begin{tabular}{|p{3.5cm}|llll|}
\hline
\multicolumn{2}{|l|}{\textbf{ALL FEATURES}} & \multicolumn{3}{|l|}{\textbf{~~~~~~~~~~~~~~~~~~~~~Train}}                                                    \\ \hline
\multicolumn{1}{|c|}{\multirow{4}{*}{\textbf{Test}}} & \multicolumn{1}{p{2cm}|}{} & \multicolumn{1}{p{2cm}|}{\textbf{~~~~~~All}} & \multicolumn{1}{p{2cm}|}{\textbf{~~~~White}} & \multicolumn{1}{p{2cm}|}{\textbf{~~~~Black}} \\ \cline{2-5} 
                  & \multicolumn{1}{l|}{\textbf{All}} & \multicolumn{1}{|c|}{65\%} & \multicolumn{1}{|c|}{65\%} & \multicolumn{1}{|c|}{60\%} \\ \cline{2-5} 
                  & \multicolumn{1}{l|}{\textbf{White}} & \multicolumn{1}{|c|}{68\%} & \multicolumn{1}{|c|}{67\%} & \multicolumn{1}{|c|}{60\%} \\ \cline{2-5} 
                  & \multicolumn{1}{l|}{\textbf{Black}} & \multicolumn{1}{|c|}{57\%} & \multicolumn{1}{|c|}{58\%} & \multicolumn{1}{|c|}{59\%} \\ \hline
\end{tabular}

\vspace{0.2cm}

\begin{tabular}{|p{3.5cm}|llll|}
\hline
\multicolumn{2}{|l|}{\textbf{WHOLE BREAST}} & \multicolumn{3}{|l|}{\textbf{~~~~~~~~~~~~~~~~~~~~~Train}}                                                    \\ \hline
\multicolumn{1}{|c|}{\multirow{4}{*}{\textbf{Test}}} & \multicolumn{1}{p{2cm}|}{} & \multicolumn{1}{p{2cm}|}{\textbf{~~~~~~All}} & \multicolumn{1}{p{2cm}|}{\textbf{~~~~White}} & \multicolumn{1}{p{2cm}|}{\textbf{~~~~Black}} \\ \cline{2-5} 
                  & \multicolumn{1}{l|}{\textbf{All}} & \multicolumn{1}{|c|}{61\%} & \multicolumn{1}{|c|}{62\%} & \multicolumn{1}{|c|}{53\%} \\ \cline{2-5} 
                  & \multicolumn{1}{l|}{\textbf{White}} & \multicolumn{1}{|c|}{62\%} & \multicolumn{1}{|c|}{63\%} & \multicolumn{1}{|c|}{51\%} \\ \cline{2-5} 
                  & \multicolumn{1}{l|}{\textbf{Black}} & \multicolumn{1}{|c|}{57\%} & \multicolumn{1}{|c|}{57\%} & \multicolumn{1}{|c|}{61\%} \\ \hline
\end{tabular}

\vspace{0.2cm}

\begin{tabular}{|p{3.5cm}|llll|}
\hline
\multicolumn{2}{|l|}{\textbf{FGT}} & \multicolumn{3}{|l|}{\textbf{~~~~~~~~~~~~~~~~~~~~~Train}}                                                    \\ \hline
\multicolumn{1}{|c|}{\multirow{4}{*}{\textbf{Test}}} & \multicolumn{1}{p{2cm}|}{} & \multicolumn{1}{p{2cm}|}{\textbf{~~~~~~All}} & \multicolumn{1}{p{2cm}|}{\textbf{~~~~White}} & \multicolumn{1}{p{2cm}|}{\textbf{~~~~Black}} \\ \cline{2-5} 
                  & \multicolumn{1}{l|}{\textbf{All}} & \multicolumn{1}{|c|}{67\%} & \multicolumn{1}{|c|}{64\%} & \multicolumn{1}{|c|}{61\%} \\ \cline{2-5} 
                  & \multicolumn{1}{l|}{\textbf{White}} & \multicolumn{1}{|c|}{68\%} & \multicolumn{1}{|c|}{67\%} & \multicolumn{1}{|c|}{60\%} \\ \cline{2-5} 
                  & \multicolumn{1}{l|}{\textbf{Black}} & \multicolumn{1}{|c|}{62\%} & \multicolumn{1}{|c|}{56\%} & \multicolumn{1}{|c|}{62\%} \\  \hline
\end{tabular}

\vspace{0.2cm}

\begin{tabular}{|p{3.5cm}|llll|}
\hline
\multicolumn{2}{|l|}{\textbf{TUMOUR}} & \multicolumn{3}{|l|}{\textbf{~~~~~~~~~~~~~~~~~~~~~Train}}                                                    \\ \hline
\multicolumn{1}{|c|}{\multirow{4}{*}{\textbf{Test}}} & \multicolumn{1}{p{2cm}|}{} & \multicolumn{1}{p{2cm}|}{\textbf{~~~~~~All}} & \multicolumn{1}{p{2cm}|}{\textbf{~~~~White}} & \multicolumn{1}{p{2cm}|}{\textbf{~~~~Black}} \\ \cline{2-5} 
                  & \multicolumn{1}{l|}{\textbf{All}} & \multicolumn{1}{|c|}{67\%} & \multicolumn{1}{|c|}{65\%} & \multicolumn{1}{|c|}{59\%} \\ \cline{2-5} 
                  & \multicolumn{1}{l|}{\textbf{White}} & \multicolumn{1}{|c|}{68\%} & \multicolumn{1}{|c|}{67\%} & \multicolumn{1}{|c|}{58\%} \\ \cline{2-5} 
                  & \multicolumn{1}{l|}{\textbf{Black}} & \multicolumn{1}{|c|}{65\%} & \multicolumn{1}{|c|}{57\%} & \multicolumn{1}{|c|}{61\%} \\ \hline
\end{tabular}

\caption{Tumour molecular subtype classification accuracy for \emph{Luminal A} vs. \emph{non-Luminal A} task. Results presented as percentage accuracy and reported for training/testing using all subjects, White subjects only and Black subjects only. From top-to-bottom: results computed using all radiomics features, just whole breast features, just fibroglandular tissue (FGT) features and just tumour features.}
\label{table:luminalclass}
\end{center}
\end{table}

\subsection{Covariate Analysis}

Next we investigated a range of covariates to test for the presence of confounding variable(s) that could be leading to the observed bias. From the full set of patient data available within the dataset we selected those variables that could most plausibly have associations with both race and model performance. These variables are summarised in Table \ref{table:covariates}. For the continuous variable (i.e. age), the table shows the median and lower/upper quartiles for White and Black patients separately. For categorical variables (i.e. all other variables), counts and percentages are provided. The $p$-values were computed using a Mann-Whitney U test for age and Chi-square tests for independence for all other variables. We can see that three of the covariates showed significant differences (at 0.05 significance) in their distributions between White and Black subjects: age, estrogen receptor status and neoadjuvant chemotherapy.

As stated earlier, non-luminal breast cancer, which is generally estrogen receptor negative, is more commonly seen in Black subjects than White subjects \cite{AbdElRehim2004,Jones2022}. In addition, this cancer is more commonly treated with neoadjuvant chemotherapy, whereas luminal breast cancer is treated with surgery, followed by endocrine therapy and chemotherapy \cite{Uchida2013} \cite{Domergue2022}. This may contribute to the statistically significant differences seen in the covariates.

\begin{table}[h]
\begin{center}
\begin{tabular}{|ll|l|l|l|}
\hline
\multicolumn{2}{|l|}{\textbf{Covariate}} & \multicolumn{1}{|c|}{\textbf{White}} & \multicolumn{1}{|c|}{\textbf{Black}} & \multicolumn{1}{|c|}{\textbf{$p$-value}} \\ \hline
\multicolumn{2}{|l|}{Age at diagnosis (years, M(L,U))}            & \multicolumn{1}{|c|}{$53.3 (45.9, 61.8)$} & \multicolumn{1}{|c|}{$50.5 (44.0, 58.5)$} & \multicolumn{1}{|c|}{$0.012$} \\ \hline
 Scanner (N / \%):    & GE       & \multicolumn{1}{|c|}{451 / 69.3\%} & \multicolumn{1}{|c|}{134/ 66.0\%} & \multicolumn{1}{c|}{\multirow{3}{*}{0.430}} \\ \cline{3-4} 
                          & Siemens  & \multicolumn{1}{|c|}{200 / 30.7\%} & \multicolumn{1}{|c|}{69 / 34.0\%} &  \\ \hline
 Field strength (N / \%):    & 1.5T       & \multicolumn{1}{|c|}{315 / 48.4\%} & \multicolumn{1}{|c|}{111 / 54.7\%} & \multicolumn{1}{c|}{\multirow{3}{*}{0.258}} \\ \cline{3-4} 
                          & 2.89T  & \multicolumn{1}{|c|}{1 / 0.1\%} & \multicolumn{1}{|c|}{0 / 0.0\%} &  \\ \cline{3-4}
                          & 3T  & \multicolumn{1}{|c|}{335/ 51.5\%} & \multicolumn{1}{|c|}{92 / 45.3\%} &  \\ \hline
 Menopause at diagnosis   & Pre       & \multicolumn{1}{|c|}{276 / 42.4\%} & \multicolumn{1}{|c|}{94 / 46.3\%} & \multicolumn{1}{c|}{\multirow{3}{*}{0.574}} \\ \cline{3-4} 
  (N / \%):               & Post  & \multicolumn{1}{|c|}{364 / 55.9\%} & \multicolumn{1}{|c|}{105 / 51.7\%} &  \\ \cline{3-4}
                          & N/A & \multicolumn{1}{|c|}{11 / 1.7\%} & \multicolumn{1}{|c|}{4 / 2.0\%} &  \\ \hline
 Estrogen receptor status & Positive       & \multicolumn{1}{|c|}{510/ 78.3\%} & \multicolumn{1}{|c|}{123 / 60.6\%} & \multicolumn{1}{c|}{\multirow{2}{*}{7.430e-07}} \\ \cline{3-4} 
  (N / \%):               & Negative & \multicolumn{1}{|c|}{141 / 21.7\%} & \multicolumn{1}{|c|}{80 / 39.4\%} &  \\ \hline
  Human epidermal growth  & Positive       & \multicolumn{1}{|c|}{111 / 17.1\%} & \multicolumn{1}{|c|}{36 / 17.7\%} & \multicolumn{1}{c|}{\multirow{2}{*}{0.906}} \\ \cline{3-4} 
  factor 2 receptor status  (N / \%):    & Negative & \multicolumn{1}{|c|}{540 / 82.9\%} & \multicolumn{1}{|c|}{167 / 82.3\%} &  \\ \hline
   Adjuvant radiation     & Yes       & \multicolumn{1}{|c|}{434 / 67.7\%} & \multicolumn{1}{|c|}{144 / 71.0\%} & \multicolumn{1}{c|}{\multirow{2}{*}{0.341}} \\ \cline{3-4} 
   therapy (N / \%):  & No & \multicolumn{1}{|c|}{210 / 32.3\%} & \multicolumn{1}{|c|}{58 / 29.0\%} &  \\ \hline
    Neoadjuvant radiation     & Yes       & \multicolumn{1}{|c|}{13 / 2.0\%} & \multicolumn{1}{|c|}{7/ 3.4\%} & \multicolumn{1}{c|}{\multirow{2}{*}{0.358}} \\ \cline{3-4} 
    therapy (N / \%):  & No & \multicolumn{1}{|c|}{632 / 98.0\%} & \multicolumn{1}{|c|}{7 / 96.6\%} &  \\ \hline
    Adjuvant chemotherapy & Yes       & \multicolumn{1}{|c|}{391/ 63.1\%} & \multicolumn{1}{|c|}{108 / 57.1\%} & \multicolumn{1}{c|}{\multirow{2}{*}{0.167}} \\ \cline{3-4} 
     (N / \%):            & No & \multicolumn{1}{|c|}{229 / 36.9\%} & \multicolumn{1}{|c|}{81 / 42.9\%} &  \\ \hline
    Neoadjuvant chemotherapy & Yes       & \multicolumn{1}{|c|}{178/ 28.1\%} & \multicolumn{1}{|c|}{91 / 46.9\%} & \multicolumn{1}{c|}{\multirow{2}{*}{1.593e-06}} \\ \cline{3-4} 
     (N / \%):                & No & \multicolumn{1}{|c|}{455 /71.9 \%} & \multicolumn{1}{|c|}{103 / 53.1\%} &  \\ \hline
\end{tabular}
\caption{Distributions of covariates in the dataset by race (White and Black subjects only). Continuous variables are reported as median (M), lower (L) and upper (U) quartiles. Categorical variables are reported as count (N) and percentage (\%). $p$-values calculated using Mann Whitney U tests for continuous variables and Chi Square tests for independence for categorical variables.}
\label{table:covariates}
\end{center}
\end{table}

\section{Discussion and Conclusions}

The main contribution of this paper has been to present the first investigation focused on potential bias in AI models trained using radiomics features. The work described in \cite{Saha2018} also reported performance of their AI models based on radiomics features broken down by race. However, in our work we have performed a more controlled analysis to investigate the potential for bias and its possible causes. As a second key contribution, our paper represents the first investigation into bias in AI models based on breast DCE-MRI imaging.

Our key findings are that: (i) radiomics features derived from breast DCE-MRI data contain race-identifiable information, leading to the potential for bias in AI models trained using such data, and (ii) RF models trained to predict tumour molecular subtype seem to exhibit biased behaviour when trained using race-imbalanced training data.

These findings show that the process of producing hand-crafted features such as radiomics features does not \emph{remove} the potential for bias from the imaging data, and so further investigation of the performances of other similar models is warranted. However, an unanswered question is whether the production of hand-crafted features \emph{reduces} the potential for bias. To investigate this, in future work we will compare bias in radiomics-based AI models to similar image-based AI models.

Our analysis of covariates did highlight several possible confounders, so we emphasise that the cause of the bias we have observed remains to be established. In future work we will perform further analysis of these potential confounders, including of interactions between multiple variables, to help determine this cause.

Interestingly, the work described in \cite{Saha2018}, which included the same \emph{Luminal A} vs. \emph{non-Luminal A} classification task using the same dataset did not find a statistically significant difference in performance between races. However, there are a number of differences between our work and \cite{Saha2018}. First, \cite{Saha2018} used all training data (half of the full dataset) when training their RF models, i.e. they did not create deliberately imbalanced training sets as we did. Therefore, their race distribution was presumably similar to that of the full dataset (i.e. 70\% White, 22\% Black, 8\% other races). It may be that this was not a sufficient level of imbalance to result in biased performances, and/or that the presence of other races (apart from White and Black) in the training and test sets reduced the bias effect. Second, we also note that the comparison performed in \cite{Saha2018} was between White and other races, whereas we compared White and Black races. Third, in \cite{Saha2018} a feature selection step was employed to optimise performance of their models. It is possible that this reduced the potential for bias by removing features that contained race-specific information, although our race classification results (see Section \ref{section:raceclassification}) suggest that this information is present across all categories of feature.

In this work we have focused on distributional shift in imaging data (and derived features) as a cause of bias, but bias can also arise from other sources, such as bias in data acquisition, annotations, and use of the models after deployment \cite{McCradden2020,Chen2021}. We emphasise that by focusing on this specific cause of bias we do not believe that others should be neglected, and we argue for the importance of considering possible bias in all parts of the healthcare AI pipeline.

Finally, this paper has focused on highlighting the \emph{presence} of bias, and we have not addressed the important issue of what should be \emph{done} about it. Bias mitigation techniques have been proposed and investigated in a range of medical imaging problems \cite{PuyolAnton2021,Zhang2022,Zong2023}, and approaches such as these may have a role to play in addressing the bias we have uncovered in this work. However, when attempting to mitigate bias one should bear in mind that the classification tasks of different protected groups may have different levels of difficulty, making it challenging to eliminate bias completely. Furthermore, one should take care to ensure that the performances of the protected groups are `levelled up' rather than `levelled down' \cite{Petersen2023} to avoid causing harm to some protected groups.

\subsubsection{Acknowledgements}

This work was supported by the National Institute
for Health Research (NIHR) Biomedical Research Centre at Guy’s and
St Thomas’ NHS Foundation Trust and King’s College London, United
Kingdom. Additionally this research was funded in whole, or in part,
by the Wellcome Trust, United Kingdom WT203148/Z/16/Z. The views expressed in this paper are those of the authors and not necessarily those of the NHS, the NIHR or the Department of Health and Social Care.

\bibliographystyle{splncs04}
\bibliography{references}

\begin{thebibliography}{10}
\providecommand{\url}[1]{\texttt{#1}}
\providecommand{\urlprefix}{URL }
\providecommand{\doi}[1]{https://doi.org/#1}

\bibitem{Abbasi-Sureshjani2020}
Abbasi-Sureshjani, S., Raumanns, R., Michels, B.E.J., Schouten, G., Cheplygina,
  V.: Risk of training diagnostic algorithms on data with demographic bias. In:
  Proceedings of {MICCAI} Workshop on Interpretable and Annotation-Efficient
  Learning for Medical Image Computing. pp. 183--192. Springer International
  Publishing (2020)

\bibitem{AbdElRehim2004}
Abd El-Rehim, D.M., Pinder, S.E., Paish, C.E., Bell, J., Blamey, R.W.,
  Robertson, J.F., Nicholson, R.I., Ellis, I.O.: Expression of luminal and
  basal cytokeratins in human breast carcinoma. The Journal of Pathology
  \textbf{203}(2),  661–671 (2004)

\bibitem{Chen2021}
Chen, I.Y., Pierson, E., Rose, S., Joshi, S., Ferryman, K., Ghassemi, M.:
  Ethical machine learning in healthcare. Annu Rev Biomed Data Sci
  \textbf{4}(1),  123--144 (2021)

\bibitem{Domergue2022}
Domergue, C., Martin, E., Lemarié, C., Jézéquel, P., Frenel, J.S., Augereau,
  P., Campone, M., Patsouris, A.: Impact of her2 status on pathological
  response after neoadjuvant chemotherapy in early triple-negative breast
  cancer. Cancers  \textbf{14}(10), ~2509 (2022)

\bibitem{Gichoya2022}
Gichoya, J.W., Banerjee, I., Bhimireddy, A.R., Burns, J.L., Celi, L.A., Chen,
  L.C., Correa, R., Dullerud, N., Ghassemi, M., Huang, S.C., Kuo, P.C.,
  Lungren, M.P., Palmer, L.J., Price, B.J., Purkayastha, S., Pyrros, A.T.,
  Oakden-Rayner, L., Okechukwu, C., Seyyed-Kalantari, L., Trivedi, H., Wang,
  R., Zaiman, Z., Zhang, H.: {AI} recognition of patient race in medical
  imaging: a modelling study. Lancet. Digit Health  \textbf{7500}(22) (2022)

\bibitem{Guo2021}
Guo, L.N., Lee, M.S., Kassamali, B., Mita, C., E., N.V.: Bias in, bias out:
  Underreporting and underrepresentation of diverse skin types in machine
  learning research for skin cancer detection - a scoping review. J Am Acad
  Dermatol  \textbf{87}(1),  157--159 (2021)

\bibitem{Ioannou2022}
Ioannou, S., Chockler, H., Hammers, A., King, A.P.: A study of demographic bias
  in {CNN}-based brain {MR} segmentation. In: Proceedings of {MICCAI} Workshop
  on Machine Learning in Clinical Neuroimaging ({MLCN}). pp. 13--22. Springer
  Nature Switzerland (2022)

\bibitem{Jones2022}
Jones, V.C., Kruper, L., Mortimer, J., Ashing, K.T., Seewaldt, V.L.:
  Understanding drivers of the black: White breast cancer mortality gap: A call
  for more robust definitions. Cancer  \textbf{128}(14),  2695--2697 (2022)

\bibitem{Larrazabal2020}
Larrazabal, A.J., Nieto, N., Peterson, V., Milone, D., Ferrante, E.: Gender
  imbalance in medical imaging datasets produces biased classifiers for
  computer-aided diagnosis. {Proc Natl Acad Sci U S A}  \textbf{117}(23),
  12592--12594 (2020)

\bibitem{Lee2022}
Lee, T., Puyol-Ant{\'o}n, E., Ruijsink, B., Shi, M., King, A.P.: A systematic
  study of race and sex bias in {CNN}-based cardiac {MR} segmentation. In:
  Proceedings of {MICCAI} Workshop on Statistical Atlases and Computational
  Models of the Heart ({STACOM}). pp. 233--244. Springer Nature Switzerland
  (2022)

\bibitem{Martini2022}
Martini, R., Delpe, P., Chu, T.R., Arora, K., Lord, B., Verma, A., Bedi, D.,
  Karanam, B., Elhussin, I., Chen, Y., Gebregzabher, E., Oppong, J.K., Adjei,
  E.K., Jibril~Suleiman, A., Awuah, B., Muleta, M.B., Abebe, E., Kyei, I.,
  Aitpillah, F.S., Adinku, M.O., Ankomah, K., Osei-Bonsu, E.B., Chitale, D.A.,
  Bensenhaver, J.M., Nathanson, D.S., Jackson, L., Petersen, L.F., Proctor, E.,
  Stonaker, B., Gyan, K.K., Gibbs, L.D., Monojlovic, Z., Kittles, R.A., White,
  J., Yates, C.C., Manne, U., Gardner, K., Mongan, N., Cheng, E., Ginter, P.,
  Hoda, S., Elemento, O., Robine, N., Sboner, A., Carpten, J.D., Newman, L.,
  Davis, M.B.: {African Ancestry–Associated Gene Expression Profiles in
  Triple-Negative Breast Cancer Underlie Altered Tumor Biology and Clinical
  Outcome in Women of African Descent}. Cancer Discov  \textbf{12}(11),
  2530--2551 (2022)

\bibitem{McCarthy2016}
McCarthy, A.M., Keller, B.M., Pantalone, L.M., Hsieh, M.K., Synnestvedt, M.,
  Conant, E.F., Armstrong, K., Kontos, D.: Racial differences in quantitative
  measures of area and volumetric breast density. J Natl Cancer Inst
  \textbf{108}(10) (2016)

\bibitem{McCradden2020}
McCradden, M.D., Joshi, S., Mazwi, M., Anderson, J.A.: Ethical limitations of
  algorithmic fairness solutions in health care machine learning. Lancet Digit
  Health  \textbf{2}(5),  e221--e223 (2020)

\bibitem{Mehrabi2021}
Mehrabi, N., Morstatter, F., Saxena, N., Lerman, K., Galstyan, A.: A survey on
  bias and fairness in machine learning. ACM Comput Surv  \textbf{54}(6) (2021)

\bibitem{Moore2020}
Moore, J.X., Han, Y., Appleton, C., Colditz, G., Toriola, A.T.: Determinants of
  mammographic breast density by race among a large screening population. JNCI
  Cancer Spectr  \textbf{26}(4) (2020)

\bibitem{Petersen2022}
Petersen, E., Feragen, A., da~Costa~Zemsch, M.L., Henriksen, A.,
  Wiese~Christensen, O.E., Ganz, M.: Feature robustness and sex differences in
  medical imaging: A case study in {MRI}-based alzheimer's disease detection.
  In: Proceedings of Medical Image Computing and Computer Assisted Intervention
  ({MICCAI}). pp. 88--98. Springer Nature Switzerland (2022)

\bibitem{Petersen2023}
Petersen, E., Holm, S., Ganz, M., Feragen, A.: The path toward equal
  performance in medical machine learning. Patterns  \textbf{4}(7),  100790
  (2023)

\bibitem{PuyolAnton2022}
Puyol-Ant{\'o}n, E., Ruijsink, B., Mariscal~Harana, J., Piechnik, S.K.,
  Neubauer, S., Petersen, S.E., Razavi, R., Chowienczyk, P., King, A.P.:
  Fairness in cardiac magnetic resonance imaging: Assessing sex and racial bias
  in deep learning-based segmentation. Front Cardiovasc Med  \textbf{9} (2022)

\bibitem{PuyolAnton2021}
Puyol-Ant{\'o}n, E., Ruijsink, B., Piechnik, S.K., Neubauer, S., Petersen,
  S.E., Razavi, R., King, A.P.: Fairness in cardiac {MR} image analysis: An
  investigation of bias due to data imbalance in deep learning based
  segmentation. In: Proceedings of Medical Image Computing and Computer
  Assisted Intervention ({MICCAI}). pp. 413--423. Springer International
  Publishing (2021)

\bibitem{Saha2018}
Saha, A., Harowicz, M.R., Grimm, L.J., Kim, C.E., Ghate, S.V., Walsh, R.,
  Mazurowski, M.A.: A machine learning approach to radiogenomics of breast
  cancer: a study of 922 subjects and 529 {DCE-MRI} features. Br J Cancer
  \textbf{119},  508--516 (2018)

\bibitem{Seyyed-Kalantari2021}
Seyyed-Kalantari, L., Zhang, H., McDermott, M.B.A., Chen, I.Y., Ghassemi, M.:
  Underdiagnosis bias of artificial intelligence algorithms applied to chest
  radiographs in under-served patient populations. Nat Med  \textbf{27}(12),
  2176--2182 (2021)

\bibitem{Stanley2022}
Stanley, E.A.M., Wilms, M., Mouches, P., Forkert, N.D.: Fairness-related
  performance and explainability effects in deep learning models for brain
  image analysis. J Med Imaging  \textbf{9}(6),  061102 (2022)

\bibitem{Uchida2013}
Uchida, N., Suda, T., Ishiguro, K.: Effect of chemotherapy for luminal a breast
  cancer. Yonago acta medica  \textbf{56}(2),  51–56 (2013)

\bibitem{Wang2023}
Wang, R., Chaudhari, P., Davatzikos, C.: Bias in machine learning models can be
  significantly mitigated by careful training: Evidence from neuroimaging
  studies. {Proc Natl Acad Sci U S A}  \textbf{120}(6),  e2211613120 (2023)

\bibitem{Zhang2022}
Zhang, H., Dullerud, N., Roth, K., Oakden-Rayner, L., Pfohl, S., Ghassemi, M.:
  Improving the fairness of chest {X}-ray classifiers. In: Proceedings of
  Conference on Health, Inference, and Learning. pp. 204--233 (2022)

\bibitem{Zong2023}
Zong, Y., Yang, Y., Hospedales, T.: {MEDFAIR}: Benchmarking fairness for
  medical imaging. In: Proceedings of International Conference on Learning
  Representations ({ICLR}) (2023)

\end{thebibliography}

\end{document}